# 3D Printable Gradient Lattice Design for Multi-Stiffness Robotic Fingers

Siebe J. Schouten[1§], Tomas Steenman[1§], Rens File´[1], Merlijn Den Hartog[1], Aimee Sakes[2], Cosimo Della Santina[3,4], Kirsten Lussenburg[2]*, Ebrahim Shahabi[3]*

*Abstract*— Human fingers achieve exceptional dexterity and adaptability by combining structures with varying stiffness levels, from soft tissues (low) to tendons and cartilage (medium) to bones (high). This paper explores developing a robotic finger with similar multi-stiffness characteristics. Specifically, we propose using a lattice configuration, parameterized by voxel size and unit cell geometry, to optimize and achieve fine-tuned stiffness properties with high granularity. A significant advantage of this approach is the feasibility of 3D printing the designs in a single process, eliminating the need for manual assembly of elements with differing stiffness. Based on this method, we present a novel, human-like finger, and a soft gripper. We integrate the latter with a rigid manipulator and demonstrate the effectiveness in pick and place tasks.

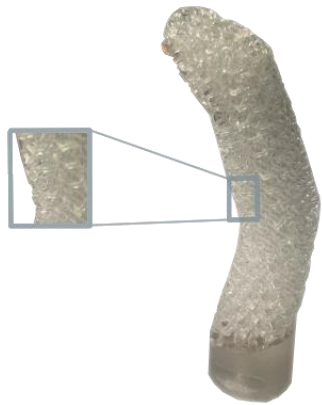

Fig. 1. An entirely 3D-printed soft robotic finger featuring an internal gradient lattice structure. The main view displays the finger's form, while the inset highlights the proposed lattice configuration, designed to vary stiffness at specific locations. This is achieved by optimizing voxel size and unit cell geometry. Varying stiffness with high granularity is a crucial step toward the long-term goal of mimicking the complex structure of the human hand. This novel finger is already capable of precise and compliant motions.

## I. INTRODUCTION

The human finger is a remarkable example of biological design. It is made up of bones, tendons, muscles, and soft tissues that work together to perform a wide range of tasks [1]. This complex design enables fingers to transition from soft tasks, such as picking up small objects, to more forceful actions, such as holding or carrying heavy objects. A key to this flexibility is changing the stiffness in different finger parts, providing flexibility and strength [2].

Soft robotics provide new solutions to solve this challenge [3], [4], [5], [6]. This technology area focuses on creating robots using highly flexible materials that can safely interact with people and adapt to different tasks [7], [8]. Soft robotic fingers provide smooth and natural movements, offering the potential to revolutionize industries such as healthcare, manufacturing, assembling, and service robotics [9], [10], [11]. Unlike rigid robotic fingers, soft robotic fingers offer enhanced dexterity, safety, and adaptability, enabling them to interact more safely with fragile objects and complex environments [12], [13], [14], [15], [16], [17]. However, soft robotics often lacks the structural stiffness required to perform functional tasks effectively. This limitation highlights a critical research gap in developing a soft robotic finger that combines sufficient structural stiffness for reliable grasping of objects.

Several researchers have addressed this challenge and proposed different solutions. Zhang et al. [18] used topology optimization to design a multi-material soft finger, improving stiffness for different applications. Matsunaga et al. [19] included a rigid component as a joint, creating a modular design for soft fingers. Zhang et al. [20]also developed a fully 3D-printed modular rigid-flexible integrated actuator for an anthropomorphic hand. Furthermore, Zang et al. [21] introduced a novel design for a multi-fingered bionic hand with variable stiffness aimed at improving robotic grasping. However, all of these studies still require the integration of several components.

Advances in manufacturing technologies have opened new possibilities for fabricating soft robotic fingers, enhancing their capabilities for complex tasks. Additive manufacturing (AM) enables the precise fabrication of complex structures that are not feasible with traditional methods. Examples are cellular materials or lattice structures [22]. These structures are complex, repeating frameworks composed of interconnected nodes and struts, forming highly organized patterns that are typically lightweight yet strong. The design of lattice structures allows for fine-tuning mechanical properties, such as stiffness, strength, and energy absorption, by altering the lattice's geometry, material, or topology [23], [24]. By carefully designing and selecting the materials for these lattices, it is possible to create components that mimic the variable

The work was in part supported under the European Union's Horizon Europe Program from Project EMERGE - Grant Agreement No. 101070918.
Siebe.J. Schouten, T.Steenman, R. File´, and M. Den Hartog are with the Mechanical Engineering Department, Delft University of Technology, Mekelweg 2, 2628 CD Delft, the Netherlands. (S.J. Schouten and T.Steenman contributed equally to this work.) A. Sakes and K.M. Lussenburg are with the BioMechanical Engineering Department, Delft University of Technology, Mekelweg 2, 2628 CD Delft, the Netherlands (Corresponding author, K.M.lussenburg@tudelft.nl). C. Della Santina and E. Shahabi are with the Cognitive Robotics Department, Delft University of Technology, Mekelweg 2, 2628 CD Delft, the Netherlands (Corresponding author, E.Shahabishalghouni@tudelft.nl). C. Della Santina is with the Cognitive Robotics Department, Delft University of Technology, Mekelweg 2, 2628CD Delft, the Netherlands. C. Della Santina is with the Institute of Robotics and Mechatronics, German Aerospace Center (DLR), 82234 Wesling, Germany (C.DellaSantina@tudelft.nl).

stiffness of human tissues [25]. This ability to adjust stiffness within the structure is important to develop advanced soft robotic systems that handle delicate and demanding tasks without sacrificing strength and accuracy.

This study explores the design and fabrication of a soft, printable robotic finger using lattice structures [26], [27], [28], [29]. By using flexible materials and employing voxel-based 3D printing, we address the common limitation of previous studies by enabling real-time, multi-directional stiffness control. Our design achieves this by adjusting lattice geometry and unit cell structure to create multiple stiffness levels throughout the finger, allowing it to bend in specific ways based on task requirements (see Fig. 1). Moreover, our design resolves the complexity of integrating multiple components, simplifying the construction with a single printing process.

The remainder of the paper is organized as follows: Section II details the design and fabrication of the lattice structure via 3D printing, along with simulation outcomes and test methodologies. Section III presents quantitative performance results, including soft finger movement, the performance of multi-stiffness of the lattice structure, and grasping capability with various objects. Section IV addresses design improvement, while Sections V and VI provide discussion and conclusions, respectively.

## II. Materials and Methods

### A. Anthromorphic Finger Design

A human finger, excluding the thumb, contains three phalanges and two joints with four degrees of freedom (DoF), enabling the finger to bend, extend, and make lateral movements. Three different anatomical layers can be distinguished: the external layer (skin), the bone layer (skeleton), and the intermediate layer (muscle and tendon tissues). These different layers give the finger different stiffnesses. Our goal is to design a soft printable robotic finger from a single material that is able to have multi-stiffness using a lattice structure. The finger was designed using Grasshopper software (Rhino 3D Version 8 SR7, Robert McNeel & Associates, Seattle, USA) with the plugins 'Crystallon' and 'Dendro.' A CAD model of a human finger was created using SolidWorks (Dassault Systemes 2023, France), after which it was imported into Grasshopper. In Grasshopper, the model is divided into individual segments called voxels, which determine the resolution of the lattice structure. Each voxel is populated with a single unit cell of the chosen lattice structure. Adjustable parameters include voxel size, unit cell design, and strut thickness, all of which influence the mechanical performance of the structure. To achieve variable stiffness at the joints, a gradient was applied to the strut thickness at specific locations within the structure.

The design of the finger in Fig. 2A is based on a human finger. The model is more equilateral than a natural finger to facilitate voxelization, and its symmetrical design enhances the uniform filling of unit cells. We chose to produce the finger slightly flexed to reduce internal stresses in the material. The length of the phalanges is based on those of an adult human male with a total finger length of 9 cm. The phalanges are set under 26° and 21°, reflecting a natural resting position (Fig. 2B). The finger will be actuated using a cable; therefore, a tube was created in the lattice, ending in a T-shape in the fingertip where the cable is fixated using a pin.

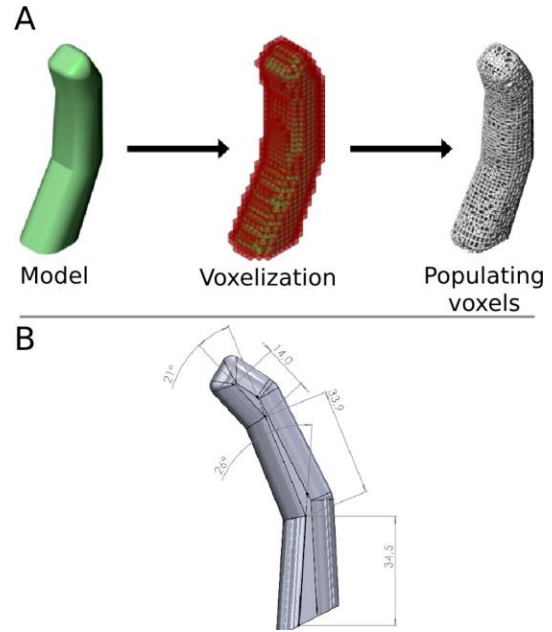

Fig. 2. Design of the soft robotic finger. (A) The soft robotic finger from model to voxelization to populating voxels. (B) The chosen geometry of the soft robotic finger.

### B. Simulation of Lattice Structure

Different unit cell designs are available in the Grasshopper software. In order to determine the unit cell design with the most suitable characteristics for the finger, we ran a simulation using COMSOL Multiphysics (COMSOL Inc, Stockholm, Sweden). Six unit cells, as shown in Fig. 3, were chosen for testing: BC, BC Cubic, Edge Octa, Vertex Octa, Trunc Octa, and Tetrahedral. To gain insight into the structural behavior of the unit cells, three different types of simulations were performed: bending, torsion, and compression. As simulating the design of the entire finger was too computationally intensive, 3 x 3 x 3 blocks of unit cells were simulated.

BioMed Elastic 50A resin V1 (Formlabs, Boston, USA) produced the finger. Young's modulus, density, and Poisson's ratio of the material are required to perform simulations. However, only the manufacturer discloses the density. The break elongation, as given by the manufacturer, was used to obtain Young's modulus. The Poisson's ratio was estimated based on similar materials with known properties [30]. The used material properties are given in Table 1.

TABLE I
Used material properties of BioMed Elastic 50A resin V1

| Property | Value |
|---|---|
| Young's Modulus | $1.8 \times 10^6$ Pa |
| Poisson's Ratio | 0.47 |
| Density | 1010 kg/m$^3$ |

TABLE II
AVERAGE STRESS AND MAXIMUM DISPLACEMENT RESULTS FROM SIMULATIONS FOR THE SIX TESTED UNIT CELLS.

|  | Average stress (Pa) | | | Maximum displacement (mm) | | |
| --- | --- | --- | --- | --- | --- | --- |
|  | Bending | Torsion | Compression | Bending | Torsion | Compression |
| **Vertex Octa** | $2.3 \times 10^4$ | $2.1 \times 10^4$ | $1.0 \times 10^4$ | 3.0 | 2.5 | 0.4 |
| **Tetrahedral** | $1.4 \times 10^4$ | $1.6 \times 10^4$ | $0.5 \times 10^4$ | 3.0 | 3.0 | 0.4 |
| **Trunc Octa** | $2.4 \times 10^4$ | $1.9 \times 10^4$ | $0.8 \times 10^4$ | 5.5 | 2.3 | 0.4 |
| **BC** | $5.3 \times 10^4$ | $3.6 \times 10^4$ | $2.5 \times 10^4$ | 10.5 | 3.5 | 1.5 |
| **BC Cubic** | $1.6 \times 10^4$ | $1.8 \times 10^4$ | $0.6 \times 10^4$ | 2.5 | 2.5 | 0.3 |
| **Edge Octa** | $2.8 \times 10^4$ | $1.9 \times 10^4$ | $1.3 \times 10^4$ | 6.0 | 2.5 | 0.9 |

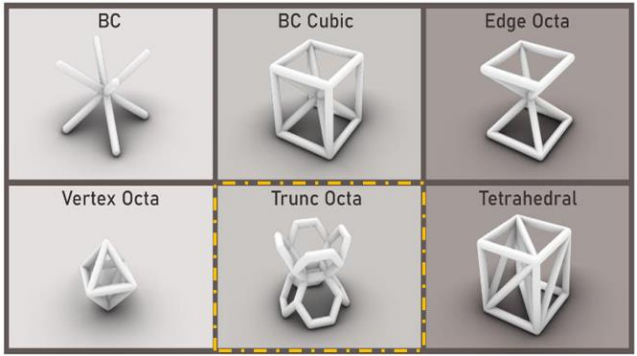

Fig. 3. Overview of six unit cell designs evaluated for their suitability in multi-stiffness soft robotic fingers.

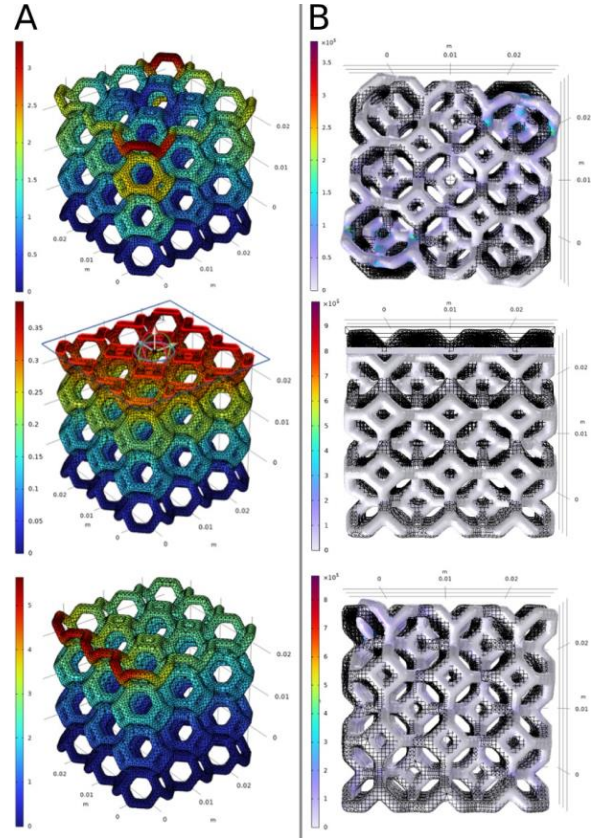

Fig. 4. Finite element analysis of a Trunc Octa lattice structure. (A) Displacement distributions show deformation patterns under applied loading conditions. (B) Stress distributions across the lattice, with variations in color representing stress concentration areas.

The displacement and stress simulations are shown in Fig. 4. and the corresponding results are given in Table 2. The red color in Fig. 4A indicates a high displacement, while the blue color indicates a low displacement. Fig. 4B shows the stress in the unit cell due to the imposed forces and gives a visual representation of the deformation of the unit cell compared to its original position. The simulation results were analyzed based on internal stresses, deflection, and average stresses. The tensile strength of the material is 2.3 MPa [30]; there should be no stresses near or over the tensile strength to prevent breakage. For deflection, it should not have too many inflections but not too few either. Thus, adjusting the gradient later in the process still has enough influence on the stiffness. Torsional stiffness should be as high as possible to be able to pick up objects at an angle as well.

The average stress was estimated for all simulations in Table 2. It can be seen that the internal stresses of bending and torsion are all fairly close to each other. The only real difference can be seen with compression: BC Cubic and Edge Octa have significantly less tension. However, all unit cells stayed well below the 2.3 MPa limit. Trunc Octa and Edge Octa score well on deflection. Trunc Octa has the least torsional displacement, followed by Vertex Octa, BC Cubic, and Edge Octa. Tetrahedral has the least displacement due to compression, followed by Vertex Octa and Trunc Octa.

Considering these results, we chose to implement Trunc Octa for the design of the unit cells.

### C. Variable Stiffness Design

The functional gradient in the lattice structure is influenced by voxel size and strut thickness. Voxel size refers to the unit cell used to build the lattice, and symmetrical voxelization is crucial—if the voxel size is mismatched, outliers occur, making the fingers less functional. A smaller voxel size reduces the functional gradient and risks fusing during printing, while larger sizes are limited by minimal lattice thickness. Two voxel sizes were tested: 2.5 mm and 4 mm.

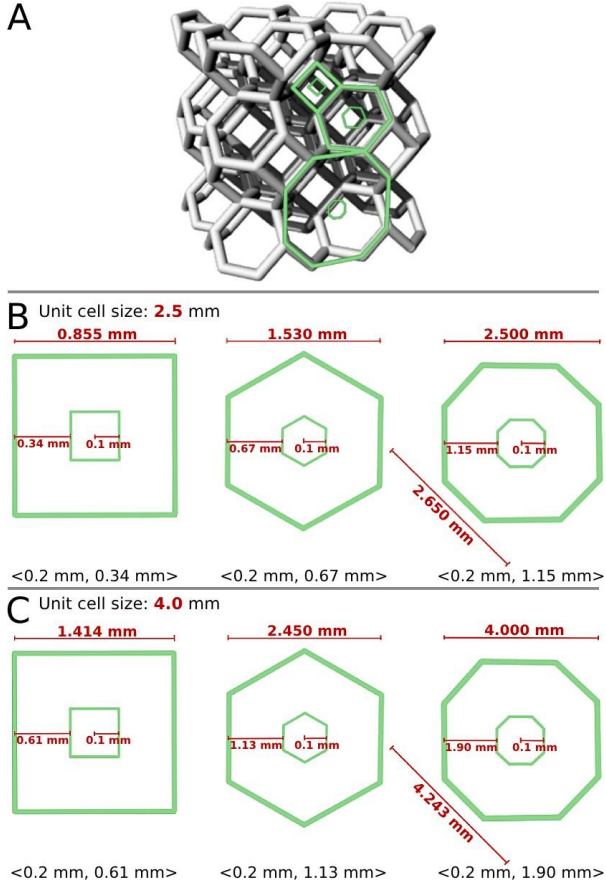

Fig. 5. Trunc Octa 2D surfaces. (A) Trunc octa 2D surfaces within the 3D structure. (B) Trunc Octa maximum thickness 2.5 mm voxels. (C) Trunc Octa maximum thickness 4 mm voxels.

The Trunc Octa unit cell's trusses impose limits on strut thickness, with a minimum gap of 0.2 mm to avoid fusing during printing. Prior prints show that a strut thickness of at least 0.2 mm is necessary for successful printing. This thickness was applied to the metacarpophalangeal joints, where maximum flexion is needed, with thicker struts farther from flexion zones to increase stiffness. Fig. 5A shows the Trunc Octa structure consists of three 2D surface types: square, hexagon, and octagon. For voxel sizes of 2.5 mm and 4 mm, the strut thickness range for each surface was determined. The lattice is still considered open if the square surface becomes solid, as the hexagons and octagons remain open. This results in lattice radii of 0.2 mm to 0.67 mm for a 2.5 mm voxel and 0.2 mm to 1.13 mm for a 4 mm voxel.

TABLE III
VOXEL SIZE AND STRUT THICKNESS PARAMETERS.

| Voxel size | Minimum strut thickness | Maximum strut thickness |
|---|---|---|
| 2.5 mm | 0.2 mm | 0.50 mm |
| 2.5 mm | 0.2 mm | 0.67 mm |
| 4.0 mm | 0.2 mm | 0.65 mm |
| 4.0 mm | 0.2 mm | 0.90 mm |
| 4.0 mm | 0.2 mm | 1.13 mm |

### D. Motion and Stiffness Test

For Motion testing, 5 fingers were 3D printed supportless, with various gradients and 2 possible voxel sizes, as shown in Table 3. Video analysis was used to compare the motion of 3D-printed fingers to a real finger, as shown in Fig. 6A. The movement was tracked using "Tracker (6.1.7)" software (Open Source Physics). Videos of both printed and real fingers, with a reference point of known size, were recorded. An XY axis was plotted in the software to set an origin, and the center of the distal metacarpophalangeal joint, which moves the most, was selected as the tracking point. The software tracked its position over time, generating X-t and Y-t plots.

The six datasets were normalized to a 0-1 second timeframe for comparison. Because each video contained different data points across varying time frames, the data were interpolated into 40 equally spaced intervals, allowing for comparison. The absolute distance between points was then calculated.

The stiffness of the fingers was tested to assess the functional properties of the different gradient ranges. An experimental setup positioned the robotic finger at various flexed angles, with a load cell (LSB200, Futek, Irvine, USA) attached to the fingertip to measure force. Simultaneously, a laser displacement sensor (optoNCDT1302, Micro-Epsilon, Ortenburg, Germany) recorded the finger's deflection, as illustrated in Fig. 6B. Both sensors were placed on adjustable rigs that allowed them to be moved and rotated accordingly. The actuation cable was fixated during all tests while the load cell was moved to deflect the finger. In its relaxed position, the finger rests at an angle of approximately 15°. It was then tested at angles of 20°, 25°, 40°, and 50°, focusing on the ranges most likely to be encountered during later application testing. The maximum test angle was set at 50°, as beyond this point, the finger, when combined with other fingers in a gripper, would no longer be able to effectively grasp objects. The measured data was processed using LabVIEW 2018 (National Instruments, Austin, USA), then cleaned and analyzed to calculate the average stiffness observed during testing.

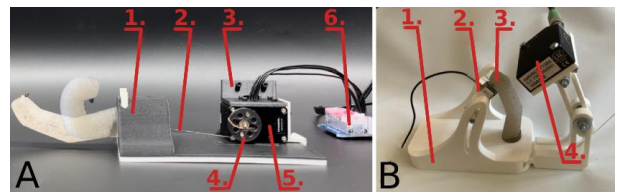

Fig. 6. Test set-ups. A) Test set-up to analyze the soft robotic finger motion. 1) Mount connecting the finger to the test setup. The finger is locked in place using two pin locks. 2) Cable for actuating finger. 3) Motor mount. 4) Pulley connecting the cable to the motor axis. 5) Motor (Dynamixel XM430-W210-R). 6) Control board. B) ) Stiffness test setup in 15° position. 1) Adjustable mount. 2)Load cell (Futek LSB200). 3) Soft finger. 4) Laser sensor (optoNCDT1302).

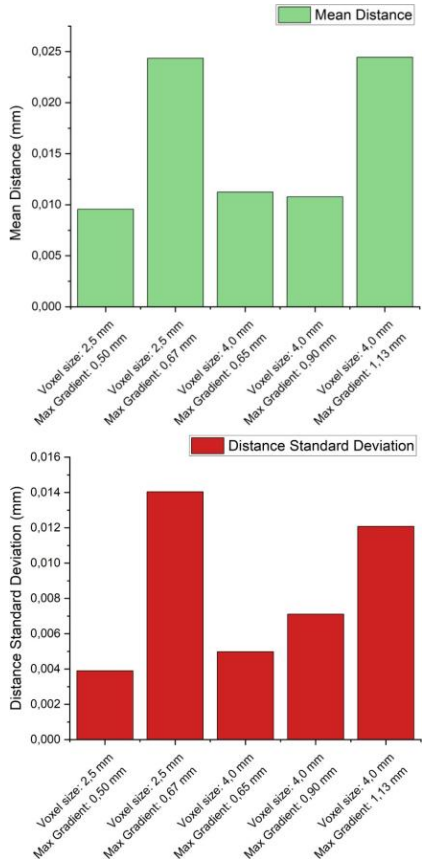
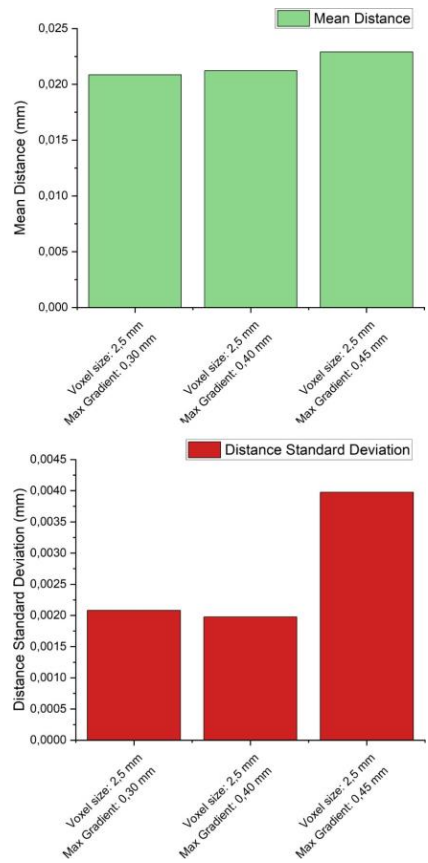

Fig. 7. Mean distances and standard deviations of tested robotic finger designs across different voxel sizes and gradient thresholds. These metrics clarify the effect of structural modifications on the accuracy and consistency of the finger's movement.

Fig. 8. Comparative analysis of mean distances and standard deviations between the tested robotic finger designs and a human finger benchmark, highlighting the degree of biomimetic achieved under different gradient configurations.

## III. RESULTS

### A. Motion results and stiffness validation

The results of the motion comparison are shown in Fig. 7. Results are reported as the mean distance and standard deviation for each comparison, providing a quantitative evaluation of the finger's movement. These results indicate that a finger with a voxel size of 2.5 mm and a gradient range of 0.2 mm-0.5 mm most accurately replicates the dynamics of a human finger, as both the mean distance and standard deviation are the smallest.

The stiffness tests were conducted exclusively on the prototype, featuring a voxel size of 2.5 mm and a gradient of 0.2 mm to 0.4 mm. The results of the stiffness test are shown in Table 4, starting with a 15° bend representing the finger in a relaxed position. The highest stiffness was measured in the 25° position, and the lowest stiffness in the relaxed position (15°). After increasing up to the 25° position, the stiffness decreased again up to the 50° position.

## IV. DESIGN IMPROVEMENT

The initial results showed that the smallest gradient with a 2.5 mm voxel size produced promising performance, particularly improving dexterity. To investigate further improvements within the lower gradient ranges, three new finger prototypes were developed, all using the same 2.5 mm voxel size. These designs were guided by the performance shown in Fig. 7, which suggests that smaller gradients contribute to greater dexterity and flexibility. The gradient ranges for the three prototypes were Finger 1: 0.2–0.30 mm, Finger 2: 0.2–0.40 mm, and Finger 3: 0.2–0.45 mm.

The performance of these newly printed fingers was evaluated with Tracker software, enabling a quantitative comparison with human finger movements. As illustrated in Fig. 8, results indicate that Fingers 1 and 2 achieved the highest performance, showing no significant difference. Both prototypes demonstrated high dexterity levels, closely mirroring human finger movement. In contrast, Finger 3 exhibited a slight decrease in precision, likely due to its broader gradient range, which may have affected its task performance. The gradient range of 0.2 mm – 0.4 mm (Finger 2) was determined to provide the best balance between dexterity and structural stability, making it particularly effective for tasks requiring flexibility.

### A. Evaluation of a Soft Finger in Real-World Scenarios

The functionality of the finger design for object grasping was evaluated using a three-fingered robotic gripper. The gripper included a 3D-printed shell that housed three actuated fingers, each powered by a servo motor (Dynamixel XM430-W210-R). This gripper was mounted on a Franka Emika Panda robotic arm, which provided six degrees of freedom, enabling precise manipulation within a three-dimensional workspace. This setup allowed for complex tasks such as picking up, manipulating, and placing objects within the robot's operational range.

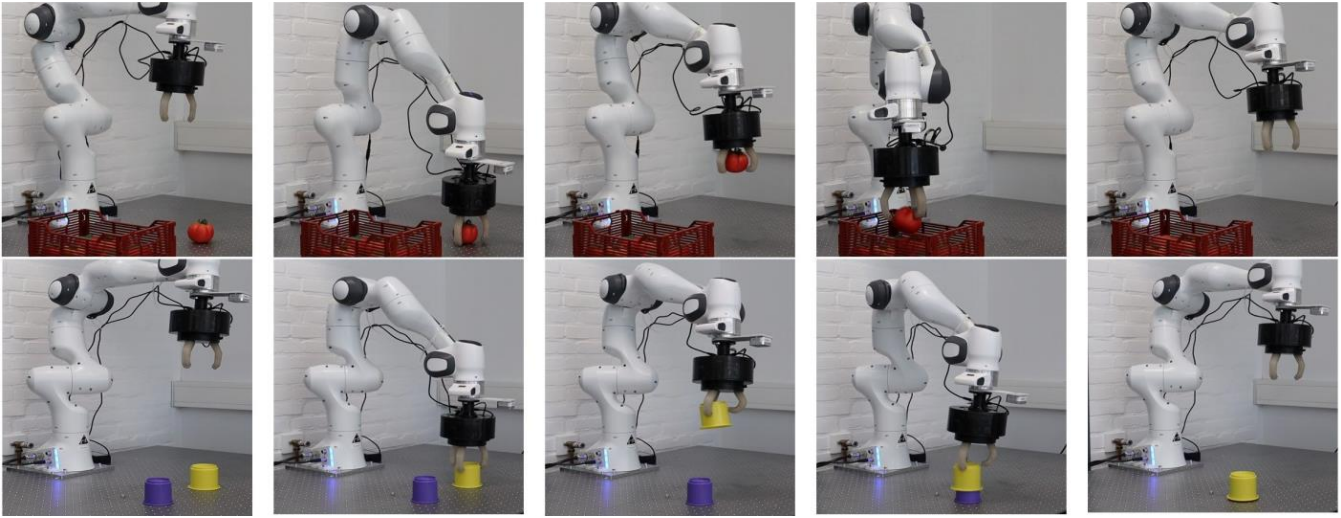

Fig. 9. Soft robotic finger demonstrating adaptability in handling an artificial tomato and a plastic mug, representing soft and rigid objects, respectively.

The gripper's performance was evaluated through a series of tests involving both soft and rigid objects (see Fig. 9). These tests were performed to assess the gripper's ability to handle varying object properties, particularly regarding flexibility and material hardness. Both sets of tests were successfully repeated multiple times, confirming the system's reliability and versatility in grasping objects with different features. The results indicate that the finger design suits various grasping tasks, supporting its potential application in various robotic manipulation scenarios.

TABLE IV
STIFFNESS MEASUREMENTS AT VARIOUS POSITIONS.

| Position [degrees] | Stiffness [N/mm] |
|---|---|
| 15° | 0.128 |
| 20° | 0.152 |
| 25° | 0.988 |
| 40° | 0.369 |
| 50° | 0.301 |

## V. DISCUSSION

The simulation and testing produced several viable finger designs. The Trunc Octa, with a voxel size of 2.5 mm and a 0.2-0.4 mm gradient, closely mimicked natural finger movement, performing best in video analysis, showing no signs of failure, and maintaining decent dexterity. Smaller voxels, in general, seemed to work better for this scale. The structure effectively manages stress concentrations, likely due to the thinner regions' reduced support from adjacent struts, contributing to failure prevention. The finger's stiffness suggests it's suitable for gripping lighter objects, though the tests only measured force on the tip with a locked cable. A full grip with active actuation might enhance strength. The stiffness closely mimics a real finger's touch but less so its overall dexterity. Noteworthy is the relatively steep stiffness changes between positions. This may be partly explained by differences in moment arm, although factors such as internal friction and practical imperfections also appear to play a role. The inherent supports of this kind of fingers proved adventitious for printing them while also leaving enough space for washing the prints before curing them. A significant limitation of this 3D printing technique is the restricted range of materials that can be used, which could constrain design flexibility. Another limitation is the constraint of printing with a single material. The Biomed material used showed great properties of flexibility and toughness. However, these materials exhibited high friction. While this was favorable for the exterior, it proved detrimental to the internal actuation tube. Especially when combined with a small contact area and relatively large forces. The finger application indicated a decent grip and object interaction, highlighting the challenges of a gripper without supplementary movements. Developing a thumb and a palm would be concrete steps to create something more constructive, like a hand. Increasing the advancements in the prosthetics industry and human-robot interactions. Short-term improvements should be made to the actuation system while further exploring the generation process. These include the implementation of asymmetrical voxels and fine-tuning the voxel size and gradient. Their interaction could result in interesting possibilities and properties. Most anticipation should come from the possibilities that single-material adjustable variable stiffness brings. Becoming a useful tool in soft robotics and even engineering in general if production techniques keep advancing.

## VI. CONCLUSIONS

In this study, a soft robotic finger with variable stiffness was developed and fabricated. The design, featuring a voxel size of 2.5 mm and a gradient thickness ranging from 0.2 mm to 0.4 mm, effectively replicates human finger movements. The finger demonstrated strong performance in object grasping and manipulation. Its flexibility and controllable stiffness make it particularly suitable for medical prosthetics and service robotics applications.

However, further improvements are necessary to broaden its potential for practical use, particularly in strength and overall stiffness. Future research directions could include the use of hybrid materials and more sophisticated 3D printing techniques to enhance the finger's robustness. Furthermore, integrating tactile sensors could provide real-time feedback during operation, enabling enhanced control precision.

## ACKNOWLEDGMENT

The authors extend their gratitude to Mariano Ramirez Montero for his invaluable assistance with setting up the Franka Emika Robot.